\def\year{2022}\relax
\documentclass[letterpaper]{article} 
\usepackage{aaai22}  
\usepackage{times}  
\usepackage{helvet}  
\usepackage{courier}  
\usepackage[hyphens]{url}  
\usepackage{hyperref}
\usepackage{graphicx} 
\usepackage{natbib}  
\usepackage{caption} 
\DeclareCaptionStyle{ruled}{labelfont=normalfont,labelsep=colon,strut=off} 
\usepackage{algorithm}
\usepackage{algorithmic}
\usepackage{amsmath,amsthm,amssymb,amsfonts}
\newtheorem{lemma}{Lemma}
\DeclareMathOperator*{\argmax}{arg\,max}

\usepackage{newfloat}
\usepackage{listings}
\floatstyle{ruled}
\newfloat{listing}{tb}{lst}{}
\floatname{listing}{Listing}
\nocopyright
\usepackage{booktabs}

\usepackage{xcolor}
\usepackage{balance}
\usepackage{multirow}

\newcommand\note{\textcolor{red}}
\newcommand\bluenote{\textcolor{blue}}

\title{Not all Failure Modes are Created Equal: \\ Training Deep Neural Networks for Explicable (Mis)Classification}
\author{
    Alberto Olmo\equalcontrib\textsuperscript{}\text{,}
    Sailik Sengupta\equalcontrib\textsuperscript{}\text{,}
    Subbarao Kambhampati\textsuperscript{}
}

\usepackage{bibentry}

\begin{document}

\maketitle
\begin{abstract}
Deep Neural Networks are often brittle on image classification tasks and known to misclassify inputs. While these misclassifications may be inevitable, all failure modes cannot be considered equal. Certain misclassifications (eg. classifying the image of a dog to an airplane) can perplex humans and result in the loss of human trust in the system. Even worse, these errors (eg. a person misclassified as a primate) can have odious societal impacts. Thus, in this work, we aim to reduce inexplicable errors. To address this challenge, we first discuss methods to obtain the class-level semantics that capture the human's expectation ($M^h$) regarding which classes are semantically close {\em vs.} ones that are far away. We show that for popular image benchmarks (like CIFAR-10, CIFAR-100, ImageNet), class-level semantics can be readily obtained by leveraging either human subject studies or publicly available human-curated knowledge bases.
Second, we propose the use of Weighted Loss Functions (WLFs) to penalize misclassifications by the weight of their inexplicability. Finally, we show that training (or fine-tuning) existing classifiers with the proposed methods lead to Deep Neural Networks that have (1) comparable top-1 accuracy, (2) more explicable failure modes on both in-distribution and out-of-distribution (OOD) test data, and (3) incur significantly less cost in the gathering of additional human labels compared to existing works.
\end{abstract}

\section{Introduction}

\looseness=-1
Deep Neural Networks have recently proven to be effective in visual classification tasks. While researchers have invested effort in trying to make these networks {\em interpretable} \cite{Expl1, Expl2, Expl3, Expl4}, we still lack a good formal understanding of how they work internally, thereby making them brittle for everyday use in real-world systems.
While mispredictions are bound to exist for any classifier that has less than cent percent accuracy, expecting a user to trust a classification system solely based on accuracy values is unreasonable. Indeed, not all failures have the same effect on a user; while some mistakes are acceptable, others can be deemed inexplicable, causing surprise and an eventual loss of human trust. Even worse, inexplicable failure modes can unwillingly step into odious stereotypes resulting in adverse societal impacts (eg. image of a dark-skinned human being misclassified as a primate \cite{google-racist}).

\begin{figure}[t]
\includegraphics[trim=0cm 0.2cm 0cm 0.5cm, width=\columnwidth]{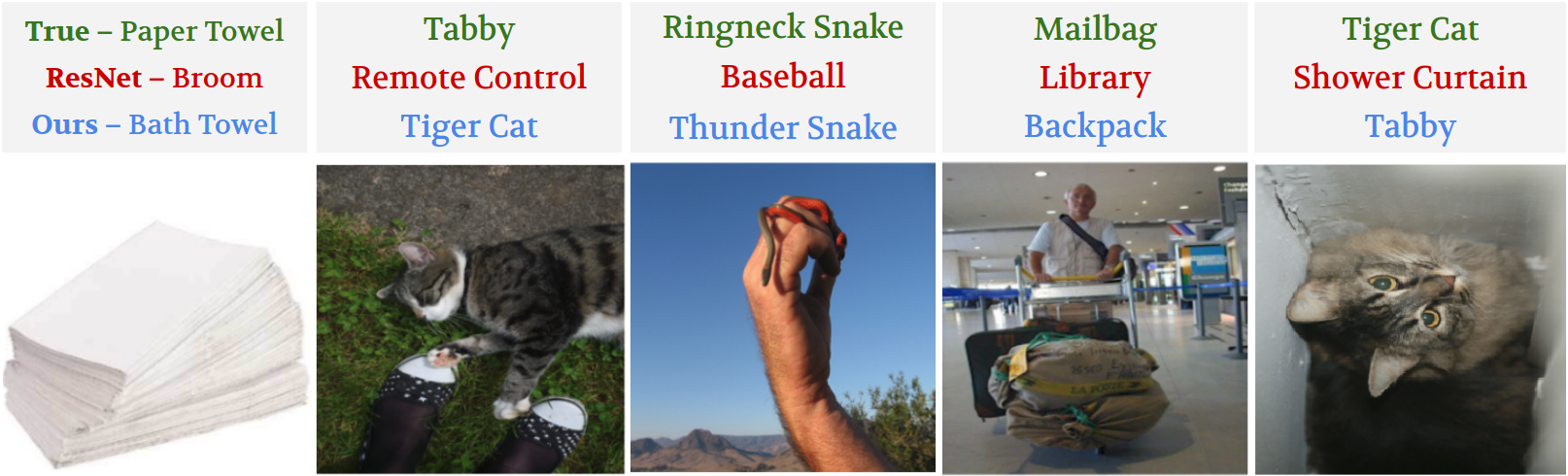}
\caption{State-of-the-art neural networks exhibit different failure modes (on ImageNet dataset), often resulting in inexplicable mistakes that can perplex users (leading to a loss of trust) and which may have unforeseen societal impacts.}
\label{fig:egregious-mistakes}
\end{figure}

We believe that egregious mistakes are a by-product of the existing loss/objective functions used by state-of-the-art classifiers; they are too sparse to encode meaningful information about failure modes. For example, the popular Categorical Cross-Entropy (CCE) loss encourages correct classification and \textit{penalizes all misclassifications equally}. In this work, we argue that incorporating the human's expectation about the failure modes ($M^h$) into the classification system ($M^r$) can help us develop explicable classifiers whose failure modes are aligned with the user's expectations.

\looseness=-1
In this regard, we answer two questions-- (1) how to represent and obtain expectations of a human (that captures the notion of egregious $vs.$ explicable misclassification), and (2) how to utilize such a representation to ensure that the trained classifier adheres to the human's expectation. To answer the first question, we posit that the notion of explicability can be represented as a semantic distance between the actual and the predicted label, i.e. misclassifications to classes semantically closer to the ground-truth are considered reasonable while misclassifications to classes further away make the end-user deem the classifier inexplicable. In particular, \textit{class-level semantic similarity can be leveraged to capture the human's expectation ($M^h$)}. To obtain $M^h$, we strongly advocate the use of a human-labeling approach, and in cases where the classification task is generic (eg. popular image classification benchmarks), we propose leveraging existing linguistic knowledge bases. Finally, to incorporate this notion of explicability into classifiers, we employ the idea of weighted loss functions to train them.

Our proposed methods demonstrate, across three image classification benchmarks, comparable accuracy and higher explicability compared to models trained with the popular categorical cross-entropy loss or prior work that advocates instance-specific human-labeling methods, which incur higher labeling costs (see \autoref{fig:egregious-mistakes}). Further, we observe in situations where classifiers are prone to making errors (eg. when faced with out-of-distribution examples), our methods consistently outperform existing baselines. Upon investigation with gradient saliency maps, we infer that our classifiers learn to focus on relevant portions of the image. Finally, we conduct a human evaluation study and show that our classifiers are indeed more explicable to users than existing ones.



\section{Related Work}

Researchers have shown that deep neural networks demonstrate inexplicable behavior in the presence of out-of-distribution \cite{NAEs,OOD1,COVIDvsCats} or adversarially perturbed test data \cite{Adv1, Adv2}, leading to a loss of human's trust in the automated system. To address these concerns, works have proposed techniques to help detect out-of-distribution \cite{OODdetect} or adversarial examples \cite{pang2018towards}. In this paper, we show that the problem is even more acute-- egregious failure modes are ubiquitous even in the context of in-distribution inputs, i.e. when the test and training distributions are similar.

The notion of explicability \cite{zhang2017plan,kulkarni2016explicable} and legibility \cite{dragan2013legibility} has been recently investigated in the context of sequential decision-making problems in task and motion planning respectively. The basic idea is that the robot performs actions using its model of the world $M^r$ and the human has an expectation about the robot's model, denoted as $M^h$. For the robot to be explicable, the authors argue that the robot should consider $M^h$ when coming up with a plan. As opposed to considering structured models to represent $M^h$, which is easier in the case of task planning scenarios \cite{kulkarni2016explicable}, we consider using labels over classification outputs to capture the human's notion of explicability in the context of computer vision tasks.

In classification tasks, existing works seek to represent the concept of \textit{trust} on black-box models in terms of the output soft-max probabilities \cite{OOD1} or the distance to the closest hyper-plane that separates the decision boundary \cite{jiang2018trust}. Other works tackle the issue of improving (a limited self-defined notion of) trust by examining a classifier's failure modes \cite{FailureModesgradCAM, FailureModesVQA}. We strongly believe that trust is difficult to define, let alone express formally, without even understanding how to represent $M^h$ or conducting human studies. Thus, our approach seeks to represent and obtain $M^h$ first. Then, we incorporate it into the classifier and finally show that it helps to prevent egregious misclassifications that can lead to loss of trust.

Our approach is similar to the idea of using soft labels as opposed to the popular notion of one-hot encoding. To understand a human's confusion about a particular test instance, works have considered interactive visual question answering \cite{SoftLabelingVisualRecognition} and obtaining humans' soft-labels for instances of a dataset \cite{CIFAR10H}. Given that the answers to instance-specific questions needs to be incorporated in training data-hungry classifiers, these approaches require enormous human effort. In contrast, we propose that $M^h$ can be done at an abstract level and tackle the problem of representing, obtaining, and incorporating $M^h$ from a class-level perspective. Note that our method thus helps to augment incomplete instance-based labeling similar to collaborative filtering \cite{sarwar2001item}.

\looseness=-1
While there exists a long history of using class-label hierarchies \cite{tousch2012semantic}, these works focus on coming up with a formal representation structure \cite{fergus2010semantic,deng2014large} or improving the speed of obtaining such representation \cite{chilton2013cascade,weld2013crowdsourcing}. On the other hand, the use of weighted loss functions (WLFs) is a common tool to penalize certain misclassifications more than others \cite{duda2012pattern,ho2019real}. For example, weighing the misclassification of inputs belonging to minorities can heavily help in soothing existing biases in data \cite{WLFAudioEventRecognition}. Similarly, using a convex loss function with weighted penalties to differentiate between \textit{quality variables} helps to find the best parameters \cite{WLFConvexLossFuncts}. We follow suit and utilize WLFs to harshly penalize the most egregious mispredictions from a human's semantic similarity perspective.


\section{Semantic Similarity}
\label{sec:ss}

\looseness=-1
In our case, semantic similarity captures the degree of inexplicability evoked in a human if a classifier were to misclassify the image of a particular class (eg. dog) to a different class (eg. cat/ship); we then want to represent this inexplicablity as the costs of misclassifications.
However, quantifying these costs is challenging given that AI systems often lack the general knowledge that humans possess. Thus, we consider three approaches, from task-specific to generic methods-- humans give instance-level costs of misclassification (task-specific), class-level costs of misclassification, and, by means of hierarchies like WordNet \cite{miller1998wordnet} we compute  the latter (generic) and obtain $M^h$. The objective is thus to obtain a pair-wise similarity metric over the classes; the distance values are proportional to the amount of inexplicability. 

\subsection{Instance-Level Human labeling (IHL)}
One way of representing the semantic similarity between the class-labels is by asking humans to label individual instances in the dataset. In doing so, we provide the human subjects knowledge about the task at hand and the available labels \cite{CIFAR10H}. This method allows one to capture a great amount of detail-- beyond (average) semantic similarity. This represents the user's expectation of explicability and also captures the robustness of $M^h$ to noise.

Unfortunately, this method suffers from two major drawbacks. First, instance-based labeling is expensive to obtain. Each image needs a significant number of humans labeling them, and data-hungry machine learning models need many such images to train. Further, as the number of class labels increases, the total number of labels required increases significantly. Second, for many tasks, there is no need for users to give labeling at such a fine-grained level.
For example, humans might find it unreasonable that the image of a dog (regardless of its breed) was misclassified to an airplane. Hence, obtaining multiple instance-specific labels renders inefficient.

In our experiments on CIFAR-10, we leverage the labels obtained via the extensive human study in \cite{CIFAR10H}. Each image was labeled by approximately 50 different people, thus having each image's label as a distribution over the classes, (i.e. soft-label) rather than just the top one. The total number of classifications for the $10,000$ images amounted to a total of $511,400$ and $2,571$ people were involved in it \cite{CIFAR10H}. We average the instance-specific human-labels over all instances of a class to obtain the semantic distance of that class to other classes.

\subsection{Class-Level Human labeling (CHL)}
In this scenario, we consider obtaining similarity labels for pair-wise class labels. For CIFAR-10, this corresponds to finding the weights on each edge of a bipartite graph matching actual class-labels to predicted ones. We gather this by performing a user study with $50$ people on Amazon Mechanical Turk \cite{turk2012amazon}.\footnote{Link to the user study: \url{https://bit.ly/3bHceX6}.} To avoid noisy answers, we only allowed participation of turkers with high reputation. Further, we added two filter questions that had trivial answers; this allowed us to detect scripted or random inputs.
Each turker was paid \$$2$ for their work that took $10$ minutes on average to complete. We gave each worker a total of $10$ questions-- one for each class label of CIFAR-10. For each class, we show a grid of $6\times 6$ sample images and ask the workers to give a score on the reasonableness of misclassifying that class w.r.t. each of the remaining $9$ classes. The answers are measured using a 0--4 Likert scale and, to avoid ambiguity, we label the 5 categories from \textit{Highly Unreasonable (surprised)} to \textit{Highly Reasonable (Explicable)}. Note the reduction in the number of human subjects required to obtain class semantic labels goes down from $2,571$ \cite{CIFAR10H} to $50$.




\subsection{Existing Knowledge for Labeling (EKL)}
Many existing image classification datasets like ImageNet \cite{deng2009imagenet}, CIFAR-10/100 \cite{CIFAR10} use nouns as class-labels. These nouns are present in popular linguistic structures such as the WordNet \cite{miller1998wordnet} (ImageNet class labels are derived from this lexical database). WordNet seeks to capture the relation between the various nouns using a tree-like structure where abstract concepts (eg. terrestrial animals) reside closer to the root and fine-grained concepts reside at the leaf of the tree (eg. Labrador retriever).

WordNet provides APIs to query various aspects of this database (eg. meaning, synonyms, antonyms, etc.). From these, \texttt{path\_similarity} returns the semantic similarity between two nodes (or words) in the database based on the hyperonym/hyponym taxonomy. We use this score for representing class-level semantic similarity. The score ranges between $[0,1]$ where $1$ means the identity mapping (i.e. comparing a class-label with itself). Given this represents a task-independent mapping, it may not be as informative for tasks that require expertise.

\section{Incorporating Semantic Similarity with Weighted Loss Functions}

Existing methods to train Deep Neural Networks encourage classification to the correct class while penalizing all misclassifications equally. The loss function objective thus, treats all failure modes indifferently, regardless of their impact on the user or the downstream task.

Weighted loss functions are often used to represent asymmetric misclassification costs for a classification task \cite{duda2012pattern}. If a task has $n$ classification labels, we consider a $n \times n$ weight matrix that encodes the different penalties when an image belonging to the ground-truth class $i$ (represented as the row) is classified to class $j$ with weight $W_{ij}$. This lets us introduce biases in the loss function to favor explicable misclassification and discourage egregious failure modes. If we represent the ground truth label as the vector $y$ with $y_i=1$ representing its membership to the actual class $i$ and the prediction vector as $p$, we can formally represent the weighted loss function for a single image, over any loss function $\mathcal{L}$, as:
\begin{equation} \label{eq:wlf}
    W\mathcal{L}F(y,p) = \mathcal{L} (W_i,p)
\end{equation}
where each row of the $W_i$ weight matrix represents the human's expectation about which misclassifications are explicable \textit{vs.} not given the actual class $i$.

\begin{figure}[t]
\centering
\includegraphics[width=\columnwidth]{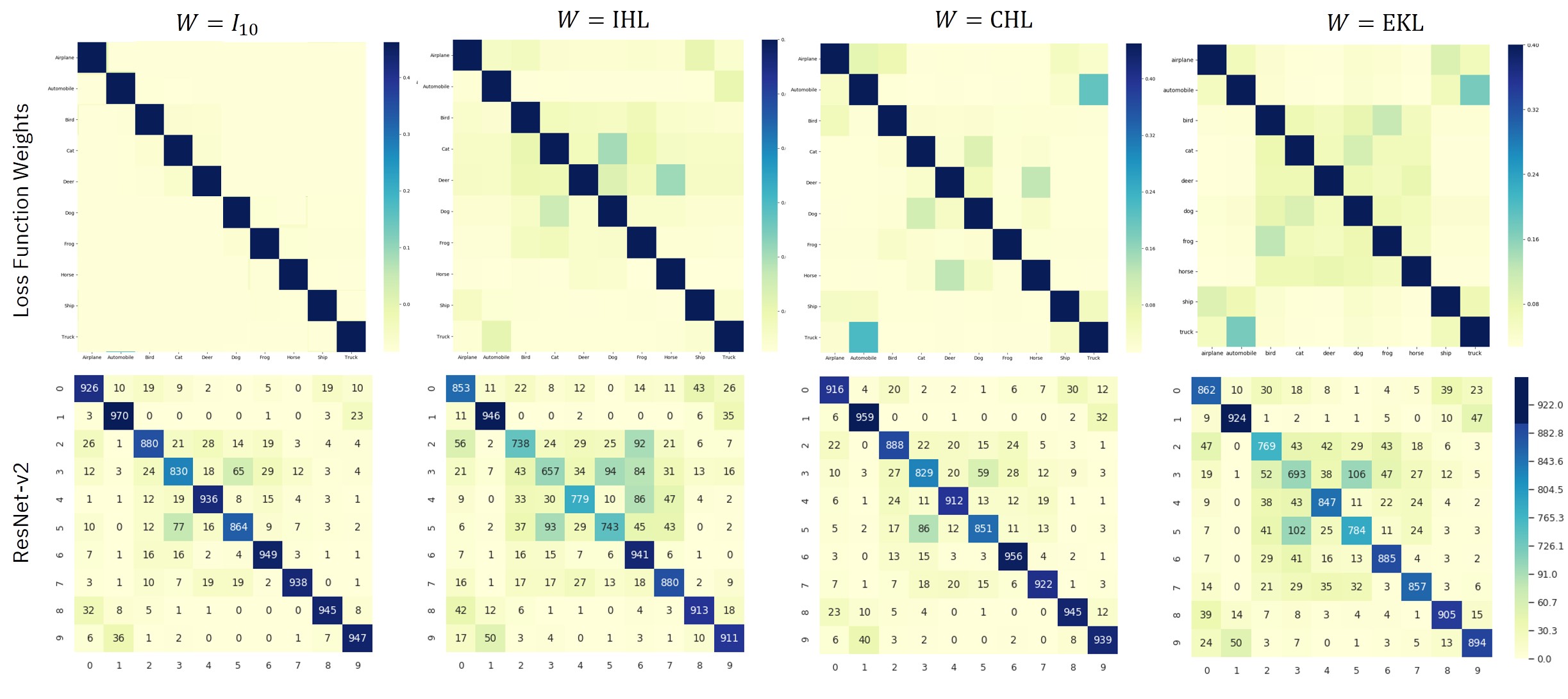}
\caption{The top row indicates the weight matrix calculated using the vanilla categorical crossentropy (where all misclassifications are weighted equally) and the three different methods (IHL, CHL and EKL). The bottom row shows that the classification results when ResNet-v2 is trained using the respective loss functions.}
\label{fig:confmats}
\end{figure}

The weight matrix contains the weights assigned to the edges of a fully connected bipartite graph from the set of actual to the set of predicted labels. In the context of the methods stated above, we add and normalize the weights provided by humans to each edge over individual instances for IHL, average the weight given to each edge by individual humans for CHL and finally, leverage distance metrics over existing knowledge bases for EKL to obtain $W$.
The different calculated weight matrices are shown in the top row of Figure \ref{fig:confmats}.
Note that the weight matrices, which capture the semantic similarity over classes for the various methods, are different. In the case of IHL \cite{CIFAR10H}, the human's uncertainty over noise in CIFAR-10 data manifests as several labels that are off-the-diagonal. In the case of EKL, every word in the lexicon is connected to every other word, and thus, we notice many (relatively) dark squares off-the-diagonal. Precisely, we notice two hierarchies-- one represented by the six classes in the middle and the other represented by the two classes at the top (or the bottom). The third column represents CHL, the human's label when faced with the question of whether a class-level misclassification is explicable or not. The deeper squares towards the bottom-left corner of $W$ represent the similarity between \textit{truck} and \textit{automobile} and have a higher weight for CHL and EKL compared to the IHL. The following lemmas show that the modified loss functions discourage egregious mistakes.

\begin{lemma}
A two-way swap between the probabilities assigned to an inexplicable vs. an explicable class penalizes the inexplicable misclassification more than the explicable one when using weighted CCEs with $W$ representing $M_h$.
\label{lem:2way}
\end{lemma}

\paragraph{Proof} In a multi-class classification problem let us assume input belongs to class $c$. Consider two incorrect classes-- an explicable misclassification label $e$ and an inexplicable misclassification label $i$. By definition $W_{ce} > W_{ci}$.

Now, we denote the classification probabilities for the two classes as $p_e$ and $p_i$. The loss function can be denoted as,
\begin{align}
    L &= - W_{ce} \log p_e - W_{ci} \log p_i + k \nonumber
\end{align}
where the third term $k$ denotes the contribution of all the other classes to the loss function. Now, let us say assume the classifier misclassifies to the the explicable class, i.e. $p_e = h > p_j ~\forall j \neq e$. Also, let $p_j = l$ ($h$ representing high and $l$ the lower probability value). Then the classifiers loss in this case is,
\begin{align}
    L_e &= - W_{ce} \log h - W_{ci} \log l + k \nonumber
\end{align}
If the probabilities of the inexplicable and explicable classes are swapped, the loss is,
\begin{align}
    L_i &= - W_{ce} \log l - W_{ci} \log h + k \nonumber
\end{align}
Now let us consider the difference between $L_e$ and $L_i$,
\begin{align}
    L_e - L_i &= \log h (W_{ci} - W_{ce}) + \log l (W_{ce} - W_{ci}) \nonumber \\
    &= (\log l - \log h) \times (W_{ce} - W_{ci}) \nonumber \\
    &\approx (-ve) \times (+ve) \nonumber \\
    &< 0 \nonumber \\
    L_e &< L_i \nonumber\\
\end{align}
where the third inequality follows from the facts that (1) $l$ and $h$ are both probability values $<1$, (2) $l < h$, and (3) log is a monotonically increasing function. As the goal of training the classifier is to minimize the loss, the lower value of $L_e$ implies, in the context of a two-way swap, the penalty incurred due to explicable misclassification is lower than the penalty incurred due to inexplicable misclassification.\hfill $\blacksquare$

\begin{lemma}
When using weighted categorical cross entropy loss with $W$ representing $M_h$, an increase $\epsilon$ in the confidence of a misclassification to class $b$ is penalized if the probability $\epsilon$ is assigned to class(es) $a$ that is(are) $\tau$-times more explicable than $b$ where, $\tau = \frac{\log ((p_a+\epsilon)/p_a)}{\log ((p_b + \epsilon)/p_b)}$, $p_a$ is the probability given to class $a$ when the misclassification to class $b$ has higher confidence, and $p_b$ is the probability given to class $b$ when the class $b$ has lower confidence.
\label{lem:theshold}
\end{lemma}

\paragraph{Proof} Let us first consider the loss structure when the class $b$ is assigned probability $p_b$ denoting the lower confidence value.
\begin{align}
    L_l &= - W_{cb} \log p_b + k_{a}^{l} \nonumber
\end{align}
where $k_a$ the log loss terms for the all other classes.

Now, let us consider that the value of $p_b = p_b+\epsilon$, i.e. the classifier classifies to class $b$ with higher confidence. In this case, the loss can be rewritten as follows,
\begin{align}
    L_h &= - W_{cb} \log (p_b + \epsilon) + k_{a}^{h} \nonumber
\end{align}
Now, we can consider the difference between the two losses.
\begin{align}
    L_h - L_l =& - W_{cb} (\log (p_b + \epsilon) - \log p_b) + (k_{a}^{h} - k_{a}^{l}) \nonumber
\end{align}
For the simplicity of the proof, let us consider the case where the difference in confidence is only assigned to a single class $a$ with probability $p_a$ when the classifier gives probability $p_b$ to class $b$ and gets $p_a + \epsilon$ when the classifier gives probability $p_b - \epsilon$ to class $b$. We now have,
\begin{align}
    L_h - L_l =& - W_{cb} (\log (p_b + \epsilon) - \log p_b) \nonumber \\
    & - W_{ca} (\log p_a - \log (p_a + \epsilon)) \nonumber \\
    =& - W_{cb} \log \frac{p_b + \epsilon}{p_b} + W_{ca} \log \frac{p_a + \epsilon}{p_a} \nonumber
\end{align}
For the classifier to penalize a misclassification to class $b$ with higher confidence more than a misclassification to class $b$ with higher confidence, we need $L_h - L_l > 0$. Thus, we have,
\begin{align}
    0 <& - W_{cb} \log \frac{p_b + \epsilon}{p_b} \nonumber \\
    & + W_{ca} \log \frac{p_a + \epsilon}{\log p_a} \nonumber \\
    \implies W_{ca} >&  \frac{\log ((p_a+\epsilon)/p_a)}{\log ((p_b + \epsilon)/p_b)} \cdot W_{cb} \nonumber \\
    \implies W_{ca} >&  ~\tau \cdot W_{cb} \nonumber
\end{align}
We first note that $p_a < p_b$ since the classifier misclassifies the image to class $b$. With the log function, the numerator has to be greater than the denominator. Hence $a$ has to be a $\tau$-times more explicable class. It is easy to note that even if the difference in confidence is distributed between more than one class, each $\tau$-times more explicable than $b$, a similar argument holds. \hfill $\blacksquare$



\section{Experimental Results}
\label{sec:expriments}

In this section, we evaluate the various classifiers on three image classification benchmarks-- CIFAR-10, CIFAR-100 and ImageNet \cite{CIFAR10,deng2009imagenet}.

\begin{table*}[]
\centering
\begin{tabular}{llccccc}
\toprule
 &  & {\em Functionality} & \em Cost & \multicolumn{2}{c}{\em Explicability} \\
 Test Data & Model & \begin{tabular}[c]{@{}c@{}}Top-1\\ Accuracy $\uparrow$\end{tabular} & \begin{tabular}[c]{@{}c@{}}Additional\\Human Labels$\downarrow$\end{tabular} & Hard$\uparrow$ & \multicolumn{1}{c}{Soft $\uparrow$}\\
 \midrule
 \multicolumn{1}{l}{CIFAR-10} & ResNet-v2 (CCE) & \textbf{91.85\%} & \textbf{0} & \textbf{18} & \multicolumn{1}{c}{\textbf{80}} \\
 & ResNet-v2 (IHL) & $83.61\%$ & +511,400 & 13 & \multicolumn{1}{c}{67}  \\
 & ResNet-v2 (CHL) & $91.17\%$ & +460 & 16 & \multicolumn{1}{c}{76} \\
 & ResNet-v2 (EKL) & $86.03\%$ & \textbf{0} & 17 & \multicolumn{1}{c}{73} \\
CIFAR-10+ & ResNet-v2 (CCE) & \textbf{48.80\%} & \textbf{0} & 780 & \multicolumn{1}{c}{2505} \\
 & ResNet-v2 (IHL) & 43.23\% & +511,400 & 703 & \multicolumn{1}{c}{2353} \\
 & ResNet-v2 (CHL) & 47.50\% & +460 & 876 & \multicolumn{1}{c}{2674} \\
 & ResNet-v2 (EKL) & 44.90\% & \textbf{0} & \textbf{1091} & \multicolumn{1}{c}{\textbf{2783}} \\
ImageNet & ResNet-v2 (CCE) & \textbf{65.20\%} & \textbf{0} & 193 & \multicolumn{1}{c}{731} \\
 & ResNet-v2 (IHL) & 55.50\% & +511,400 & 197 & \multicolumn{1}{c}{703} \\
 & ResNet-v2 (CHL) & 63.70\% & +460 & 208 & \multicolumn{1}{c}{\textbf{760}} \\
 & ResNet-v2 (EKL) & 56.71\% & \textbf{0} & \textbf{215} & \multicolumn{1}{c}{720} \\
\bottomrule
\end{tabular}
\caption{Evaluation of models trained on CIFAR-10 data.}
\label{tab:cifar10-internal}
\end{table*}

\subsection{CIFAR-10}
\label{ssec:cifar10}

\subsubsection{Metrics}
We consider three top level metrics that capture the {\em functionality}, the {\em explicability} of a classifier and the {\em additional labeling cost} associated with training that particular classifier. For functionality, we use top-1 accuracy and for cost, we consider the amount of additional labels we will need to obtain the distances described in \S3.

For explicability, prior work uses the loss function objective value as an evaluation metric \cite{CIFAR10H}. Such a metric cannot distinguish between the various methods as every classifier trained on a particular loss function (eg. $L_{CHL}$) will perform the best when evaluated using on that loss function ($L_{CHL}$) and worse-off when evaluated using the others ($L_{IHL}$, $L_{EKL}$). We conduct experiments to show this in the supplementary material. More importantly, we consider two synthetic metrics in this section to measure explicability (and conduct human studies in \S6).
To this extent, we first create a distance metric by averaging the three distances in \S3. Then, we gather all images misclassified by the set of classifiers and calculate the two measures as follows.

\noindent {\em Hard score:} is the number of times that for every misprediction $x$, class $c$ and pair of true and predicted classes, the given classifier achieved the highest similarity score amongst all classifiers' mispredictions. That is, for any given class index $c$ and intersection set of misclassification images $M$, $Hard_c=\sum_{x\in\{M\}}[1\ if\ \argmax_i(v(i))=c]$ where $v(i)$ is the similarity vector. 

\noindent {\em Soft score:} adds $1/n$ to the models whose misprediction was the least egregious where $n$ is the count of classifiers that obtained the same maximum similarity score. Hence, $Soft_c=\sum_{x\in\{M\}}[1/n\ if\ \argmax_i(v(i))=c]$.

\subsubsection{Results}

\looseness=-1
In Figure \ref{fig:confmats} we plot a heat-map showcasing the classification results obtained using ResNet-v2 trained using the three different $W$s for IHL, CHL and EKL in the bottom row. We show that the classifier effectively captures the respective semantics represented by each corresponding $W$.

In \autoref{tab:cifar10-internal}, we consider the three aforementioned metrics to evaluate classification on the CIFAR-10 test set. Note that the cost of obtaining the IHL metrics is much higher compared to the other methods. ResNet-v2 achieves the highest accuracy values and has the best soft and hard scores. As the classifiers have high accuracy, we find that only $298$ images are misclassified by all models; less samples makes the hard and soft scores unreliable to distinguish between the classifiers is terms of explicability.

\looseness=-1
Hence, we also evaluate on out-of-distribution (OOD) scenarios where the classifiers are more prone to making mistakes. On CIFAR-10+ and ImageNet data (belonging to the 10 classes in CIFAR-10), we first observe that (as expected) the top-1 accuracy becomes relatively low. We then analyze the misclassified test instances. For CIFAR-10+, we notice that ResNet-v2 (EKL) outperforms all the other classifiers in terms of the hard and soft scores, indicating higher explicability of the misclassified examples. For ImageNet, ResNet-v2 (CHL) again has the best hard score, but  loses out to ResNet-v2 (CHL) on the soft score. Furthermore, ResNet-v2 (CHL) has comparable accuracy to ResNet-v2 (CCE), the best classifier, while ResNet-v2 (EKL) lags behind by $8.49\%$.

\begin{figure}[t!]
\centering
\includegraphics[width=\columnwidth]{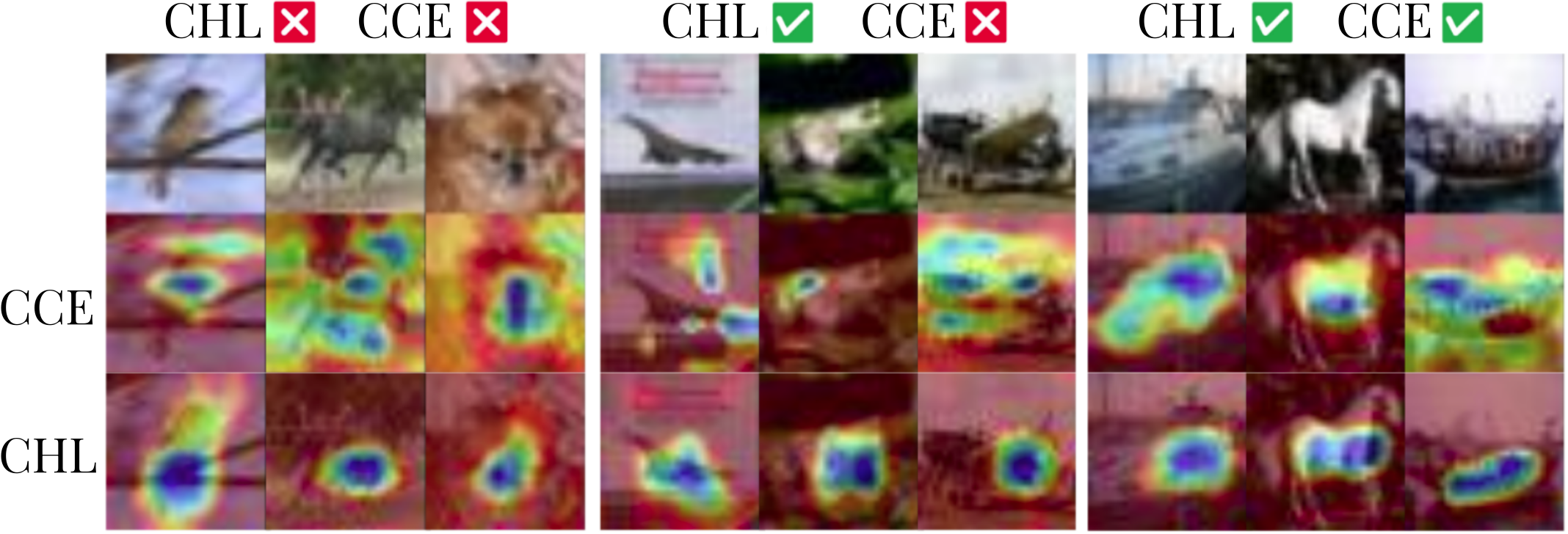}
\caption{Grad-CAM saliency maps for the ResNet-v2 (CCE) and ResNet-v2 (CHL) models. The three columns represent three cases-- (1) both base and CHL misclassify but CHL has higher explicability, (2) CHL is correct and base is not, (3) both models are correct.}
\label{fig:gradcam_table}
\end{figure}

\subsubsection{Grad-CAM attention maps} Convolutional neural networks can often classify inputs correctly with high confidence, but due to wrong reasons. To investigate this, techniques based on saliency maps, like Grad-CAM \cite{FailureModesgradCAM}, have been proposed. We use Grad-CAM to examine which input features ResNet-v2 (CHL) prioritized when making a classification decision and compare it with the Grad-CAM of ResNet-v2 (CCE). The results are shown in Figure \ref{fig:gradcam_table}.

In general, we observe that the ResNet-v2 (CCE) often focuses on regions of an image that are irrelevant for making the correct classification. For example, for the image on the second column, the ResNet-v2 (CCE) focuses on the background trees and the grass; for the image in the fourth column, the ResNet-v2 (CCE) hardly focuses on the object of interest and concentrates on the (illegible) textual writing in the image.
On the other hand, ResNet-v2 (CHL) gives more importance to the primary object in the image and hardly focuses on the background information. This observation holds even when both models are correct. In the image in the last column, note how ResNet-v2 (CHL) only focuses on the body of the ship whereas the ResNet-v2 (CCE) pays attention to the water and the buildings in the far-away horizon. We also observe an inherent bias in the CIFAR-10 data-- the object in the center of the image guides the ground truth label, which makes the ResNet-v2 (CHL) focus on the center of the images. For further experimentation on this subject, please refer to the appendix section in the supplementary material.

\begin{table}[t!]
\footnotesize
\centering
\begin{tabular}{lcccc}
\toprule
Test Data/Model & Accuracy $\uparrow$ & Hard $\uparrow$ & Soft $\uparrow$ \\
\midrule
CIFAR-100\\
\hspace{5px}VGG (CCE) & $70.48\%$ & 127 & 1373 \\
\hspace{5px}VGG (EKL) & \textbf{70.63\%} & \textbf{157} & \textbf{1403} \\
CIFAR-100+\\
\hspace{5px}VGG (CCE) & \textbf{11.70\%} & 1688 & 13066 \\
\hspace{5px}VGG (EKL) & $11.54\%$ & \textbf{1783} & \textbf{13161} \\
\bottomrule
\end{tabular}
\caption{Evaluation of models trained on CIFAR-100 data.}
\label{tab:cifar100-vgg}
\end{table}

\subsection{CIFAR-100}

CIFAR-100 contains images belonging to $100$ classes \cite{CIFAR10}. In this case, the IHL baseline requires human subjects to (1) provide a probability distribution over $100$ classes for each data-point leading to an increase in the human subject's cognitive overload, and (2) annotate a significantly larger number of labeled samples compared to CIFAR-10 increasing the cost of obtaining additional labels.
In the case of CHL, the cost of labeling, while still significantly less than IHL, also increases because we now need weights for a bipartite graph with ${100 \choose 2} = 4950$ edges. To reduce cognitive overload on the human, we can show just a subset of classes that a class can be misclassified to; this leads to an increase in the population size. Note that such a breakdown is difficult to do in the context of IHL. Owing to the added cost for both the IHL and CHL methods, we consider only EKL in this setting. Similar to the case of CIFAR-10, all the class labels present in CIFAR-100 are also a part of WordNet. Thus, we use the path similarity between the class-labels to populate the weight matrix $W$. Further, we consider fine-tuning a pre-trained model of the VGG classifier \cite{simonyan2014very} that has $\approx 183$ million parameters for this task. Thus, this experiment also helps us showcase the benefits of our approach when the classifiers are significantly larger and tasks are complex.

In \autoref{tab:cifar100-vgg}, we notice that VGG (EKL) outperforms VGG (CCE) in terms of accuracy, hard and soft scores on the in-distribution CIFAR-100 test data. In contrast to the results in the previous section, the use of a weighted loss function, which enforces a soft-labeling scheme, behaves as a regularizer and increasing the top-1 accuracy of the pre-trained VGG (CCE) from $70.48\%$ to $70.63\%$ for VGG (EKL). To reinforce our conclusion, we evaluate the models on an out-of-distribution dataset-- CIFAR-100+ (comprised of 300 additional non-overlapping CIFAR-100 images per class extracted from Flickr). Similar to \S5.1, we observe the accuracy of both the classifiers reduce. Among the misclassified examples, we notice that VGG (EKL) outperforms VGG (CCE) on the soft and hard explicability metrics.

\begin{figure}[t!]
\centering
\footnotesize
\includegraphics[width=\columnwidth]{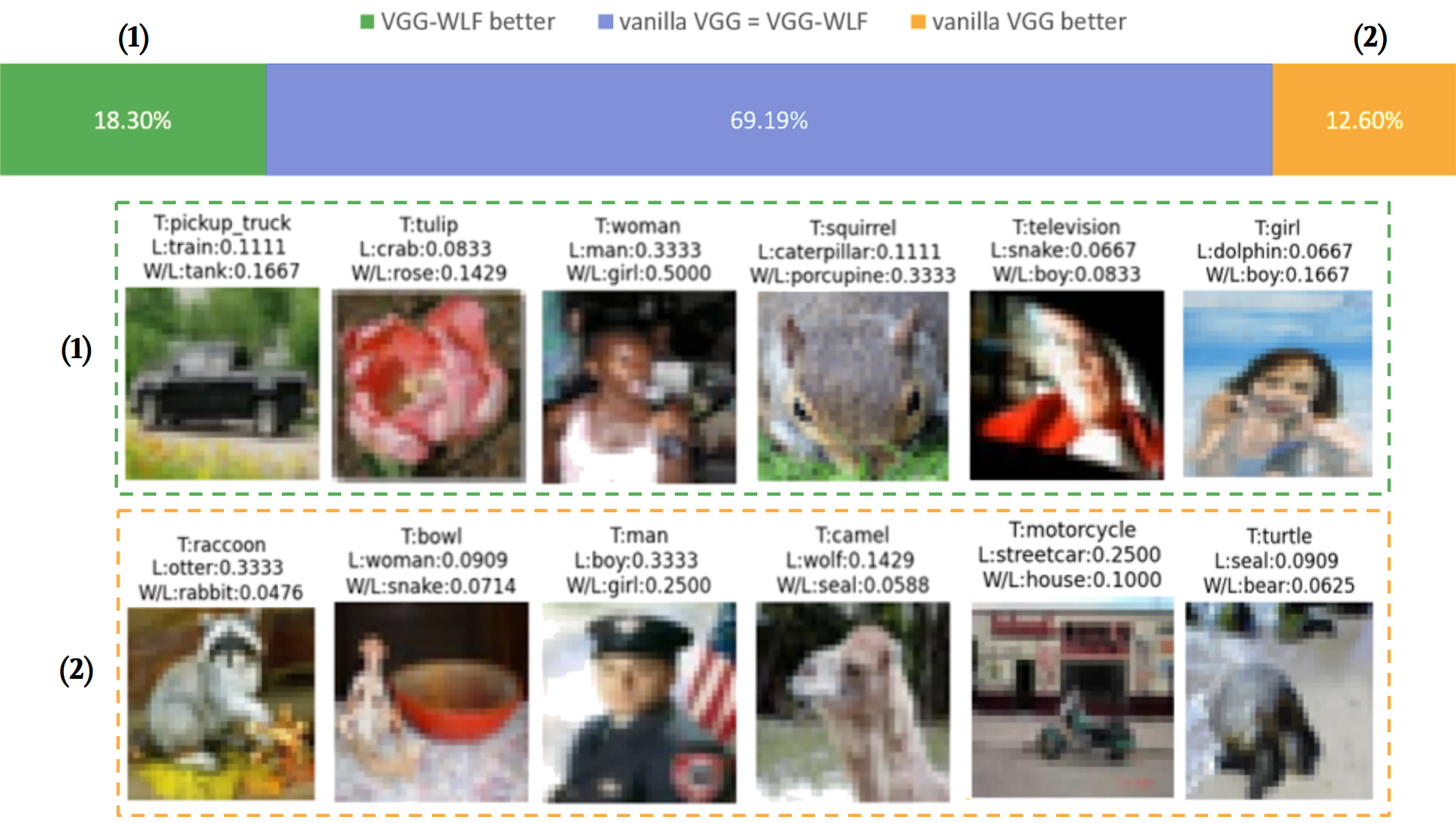}
\caption{Misclassification on CIFAR-100 test set shows that VGG-EKL makes more explicable mistakes.}
\label{fig:cf100}
\end{figure}

\looseness=-1
In Figure \ref{fig:cf100}, we showcase three scenarios that arise when both the classifiers mispredict a given test input. We show the ground-truth class label, the class label it was classified to by the vanilla VGG followed by VGG fine-tuned using EKL. The numbers beside the predicted class labels show the similarity between the predicted class and the true class as per WordNet's path similarity metric.
In the majority of the cases, precisely $69.19\%$ of them, both the classifiers misclassify an input to the same incorrect class. This high agreement should not be surprising because the VGG-EKL is simply a fine-tuned version of the vanilla VGG network. There exist two other scenarios-- (1) when VGG-WLF misclassifies an input image to a semantically closer class and (2) when the vanilla VGG does so. The former happens $18.3\%$ of the time while the latter occurs $12.6\%$ of the time. Examples of the first case show that flowers like {\tt tulip} and {\tt orchid} are classified as {\tt crabs} and {\tt plates}, an image of people are classified as animals ({\tt girl} $\rightarrow$ {\tt dolphin}), and animals are classified to inanimate objects ({\tt kangaroo}  $\rightarrow$ {\tt bottle}) by the vanilla VGG classifier. On the other hand, VGG-EKL preserves these semantics learned from WordNet.
In the latter case, examples highlight that misclassifications made by VGG-WLF, while worse-off than the vanilla VGG, are less egregious as per the Word-Net similarity metrics. This fact is reinforced by the values of the explicability metric (in Table \ref{tab:cifar100-vgg}) that is significantly better for VGG-WLF compared to vanilla VGG. 
Note that there exists a subset of test inputs on which the misclassifications made by VGG-WLF refer to an object present in an input image but, due to the original ground-truth label, regarded as a misclassification. For example, the image (in the middle) labeled as a \texttt{television} shows the picture of a person inside a television. While vanilla VGG labels it as a {\tt snake}, VGG-WLF labels it as \texttt{boy} referring to the person.

\begin{table}[t!]
\centering
\begin{tabular}{lcccc}
\toprule
Test Data/Model & Accuracy $\uparrow$ & Hard$\uparrow$ & Soft $\uparrow$\\
\midrule
ImageNet\\
\hspace{5px}Inception-v2 (CCE) & \textbf{79.68\%} &  151 & 846 \\
\hspace{5px}Inception-v2 (EKL) & 78.74\% & \textbf{183} & \textbf{878} \\
\bottomrule
\end{tabular}
\caption{Evaluations of models trained on ImageNet data.}
\label{tab:imagenet}
\end{table}

\subsection{ImageNet}

ImageNet \cite{deng2009imagenet} consists of 14 million high resolution annotated images belonging to 1000 classes. This makes gathering human-labeled instance (or even class) level confusion matrices costly. However, owing to ImageNet's class labels being a subset of WordNet's \cite{miller1998wordnet} nouns, we can leverage this semantic structure, define our EKL loss and fine-tune \textit{Inception ResNet-v2} \cite{he2016deep} for 5 epochs. In Table \ref{tab:imagenet} we show that even though there is a reduction of $0.95\%$ accuracy, EKL improves our explainability metrics in comparison. Moreover, in Table \ref{tab:human_eval} and as we discuss below, we show that it is also enough for the model to become twice as explicable as per human evaluations on the most egregious images.

\begin{table}[b!]
\begin{tabular}{llc}
\toprule
Test Data & Model & Human Rating $\uparrow$ \\ 
\midrule
ImageNet & Inception-v2 (CCE) & 1.456 \\
 & Inception-v2 (EKL) & \textbf{2.897} \\[0.2em]
CIFAR-10+ & ResNet-v2 (CCE) & 0.741 \\
 & ResNet-v2 (IHL) & \textbf{1.799} \\
 & ResNet-v2 (CHL) & 1.725 \\
 & ResNet-v2 (EKL) & 1.743 \\
 \bottomrule
\end{tabular}
\caption{Human evaluations on the egregiousness of model predictions on ImageNET and CIFAR-10+.}
\label{tab:human_eval}
\vspace{-0.5em}
\end{table}

\section{Human Evaluations}
\label{sec:human_eval}

While metric-based evaluations (in \S\ref{sec:expriments}) show that the proposed methods misclassify to semantically closer/explicable classes, we conduct additional human studies to see if the evaluation metrics align with the human's notion of explicability. We consider 50 and 60 misclassifications made by Inception-v2 on ImageNet test data and ResNet-v2 on CIFAR-10+ test data respectively. The misclassifications are chosen randomly from the set of images misclassified by all the classifiers and more than a certain threshold away from the true class as per the abovementioned confusion matrices of each classifier. Then, we present them to master workers on Mechanical Turk \cite{turk2012amazon} and provide them with 5 different options representing the Likert score that ranges from \textit{Highly Unreasonable (surprised)} to \textit{Highly Reasonable (Explicable)}. Due to the low image quality of CIFAR-10+ data, we add an extra option that lets the worker disregard the image if they are unable to make sense of it. 

The average scores across all the images for each of the models are shown in \autoref{tab:human_eval}. We can see that the classifiers trained with the weighted loss functions are, by far, more explicable to humans. For ImageNet, the human evaluation aligns with the hard and soft metrics. For CIFAR-10+, the hard and soft metrics prefer ResNet-v2 trained on EKL, whereas human evaluations prefer ResNet-v2 trained on IHL. Given the labeling cost for IHL and the low overall explicablity scores that the vanilla ResNet-v2 classifier obtains, EKL seems to be the reasonable middle ground.

\section{Conclusions}

While we talk of explicable classification, our goal is to train a classifier that agrees with a human's view of the failure modes, thereby reducing the surprise caused by a particular misclassification. A more nuanced view should consider the penalty of a mistake in terms of the various impacts a particular misclassification may have on the downstream task. In this regard, we recognize an operational perspective on misclassifications.


Often, misclassifications may be inexplicable to a human but, given the downstream task, considered reasonable. For example, in Figure \ref{fig:cf100}, classifying a {\tt kangaroo} to a {\tt bottle} may be deemed unsafe for autonomous driving scenarios (in Australia) whereas a system misclassifying it to a {\tt boy} is better as the underlying decision of stopping the car remains unaffected. Without the context of the underlying task, classifying a {\tt kangaroo} to a {\tt boy} may be considered inexplicable. Thus, the class-level penalty scores for explicability may not align with the task-specific class-level penalties for operational purposes. Leveraging existing knowledge bases, unless designed specifically for the task at hand, becomes unreasonable and in these scenarios, CHL is the only choice.

In this paper, we showed that the prevalent objective functions for training Deep Neural Networks weigh all misclassifications equally which often lead to inexplicable failure modes and loss of human trust in the system. To prevent these egregious misclassifications for vision classification tasks, we proposed two methods to obtain the human's model $M^h$ that calibrates misclassifications on a scale ranging from explicable to egregious. We note that our methods can be, beyond the explicability scenario, used to learn operational semantics or encode misclassification that have negative societal impacts. We then utilized the notion of weighted loss functions to incorporate $M^h$ into the classifier's model and showed with experiments on different datasets and models that our method indeed helps classifiers reduce the number of egregious errors.

\newpage

\bibstyle{unsrt}
\bibliography{ref}
\balance

\end{document}


\begin{figure*}[t]
    \centering
    \includegraphics[width=0.95\textwidth]{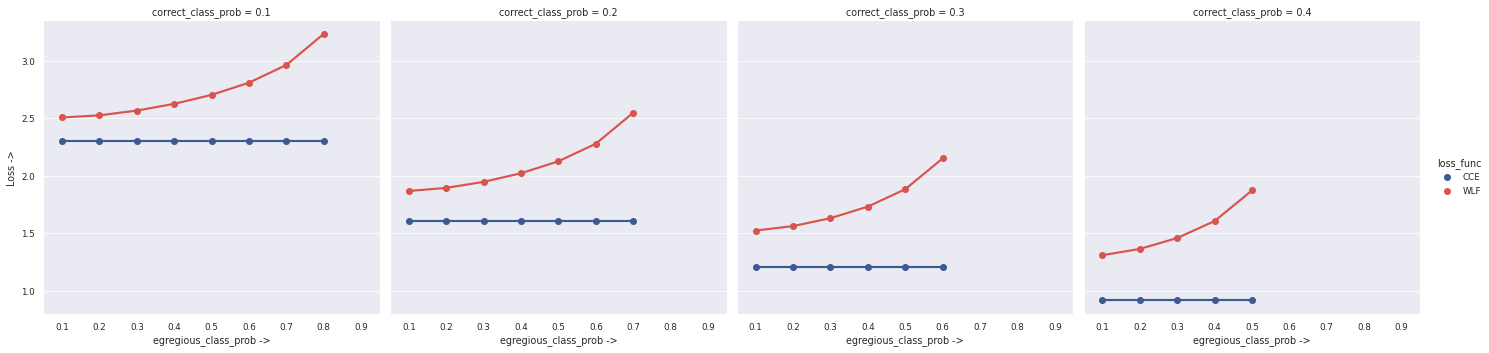}
    \includegraphics[width=0.95\textwidth]{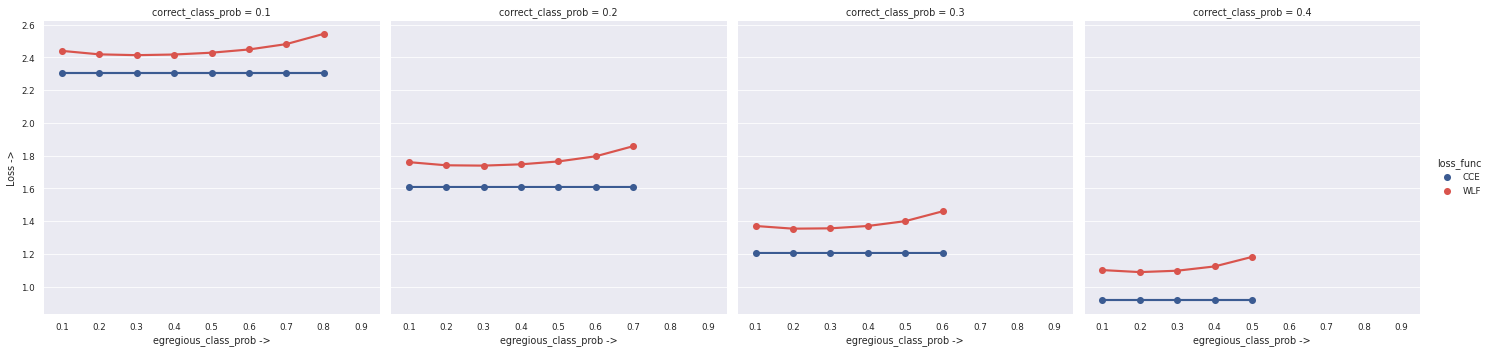}
    \includegraphics[width=0.95\textwidth]{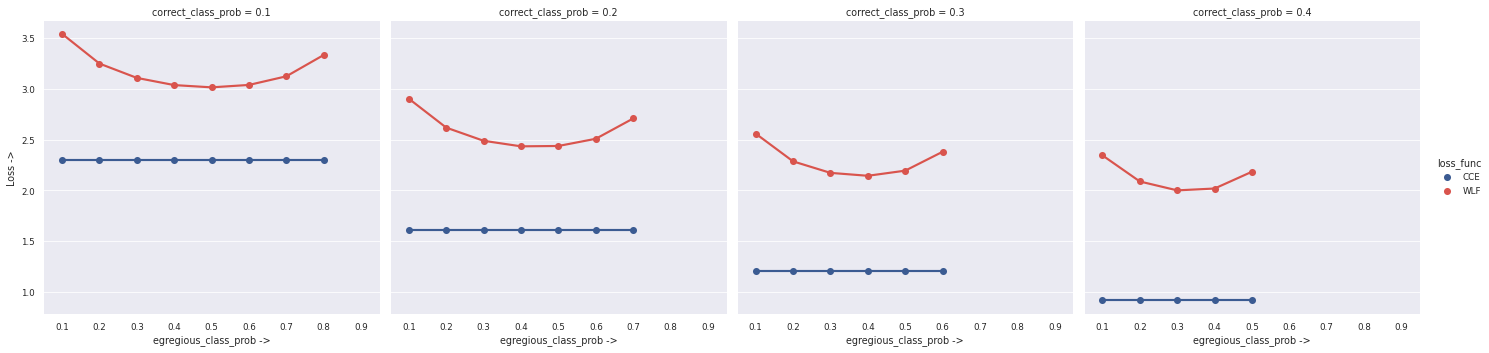}
    \caption{Comparing Categorical Cross-Entropy to our loss and characterizing the behavior of the weight CCE for a three-class classification problem when of the two misclassifications, (top) one is explicable and the other is inexplicable, (middle) when both are inexplicable, and (bottom) both are explicable.}
    \label{fig:my_label}
\end{figure*}

\section*{\centering Appendix}
We are dividing this appendix in two main sections (1) empirical validation for loss function behavior where we characterize the lemmas proposed in the paper, and (2) additional experimental results.

\section*{Empirical Validation: Characterizing the Behavior of the Proposed Loss Function with Numerical Simulations}

\looseness=-1
We empirically show that our proposed loss function, in comparison to categorical cross-entropy loss, given a richer way to capture the human's expectation over failure modes. This acts as an empirical validation of the two lemmas proved, showing the inexplicable or egregious mistakes are penalized more, especially when made with higher confidence.
For this setting, we consider a three class classification problem with the correct class, a semantically closer class with weight $W_c$ and an less semantically similar class with weight $W_f$.

\begin{figure*}[h]
    \centering
    \includegraphics[width=\textwidth]{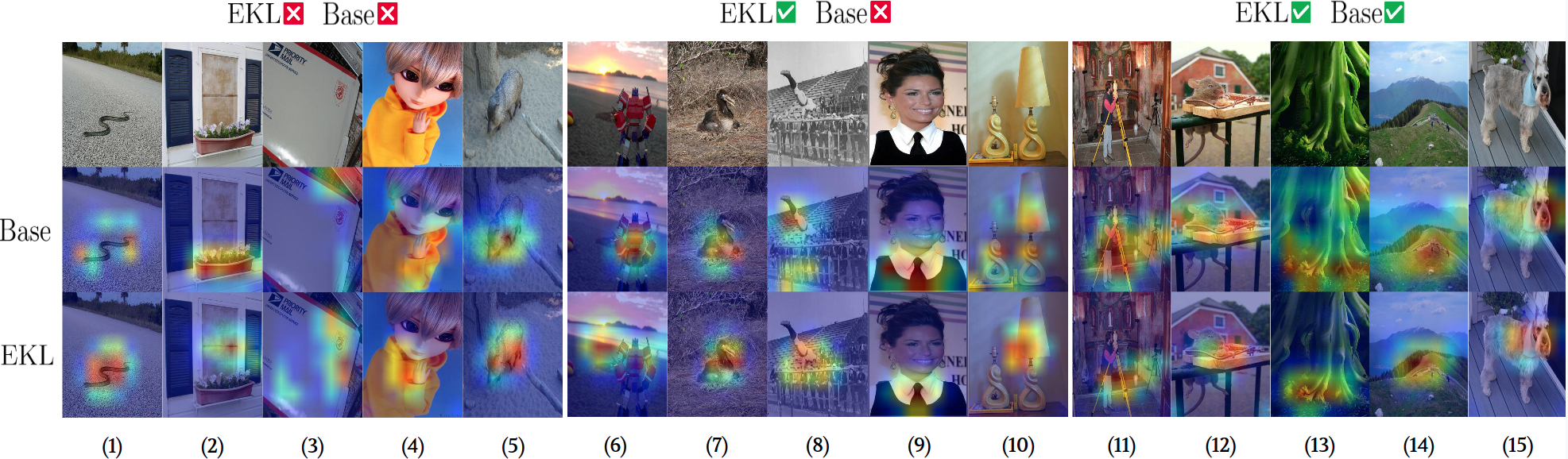}
    \caption{Grad-CAM saliency maps on ImageNet samples when (a) both our ResNet-EKL and vanilla ResNet-v2 (base) models are wrong but EKL's prediction is semantically closer (b) when ResNet-EKL's predictions are right while the base is wrong and (c) when both are correct. More calid colors (red) represent higher saliency importance.}
    \label{fig:imagenet-saliency}
\end{figure*}

\subsubsection*{When $c$ is highly explicable and $f$ is highly inexplicable}
In the setting, we consider a case where $W_c = 0.4$ and $W_f = 0.05$; this corresponds to the setting where the condition in Lemma \twowayref{} holds.
In \autoref{fig:my_label} (top), we show four graphs where the probability of the correct class increases from left to right. In each graph, we show that as the confidence on the egregious class increases along the x-axis, the value of the loss function, plotted along the y-axis, also increases, thereby highly penalizing inexplicable classification with higher confidence. Further, when the probability assigned to the correct class is higher, the magnitude of the penalty decreases significantly. Note that if the correct class if deemed to be class $a$ in Lemma \threshref, it is the easiest to satisfy the threshold $\tau$.

\subsubsection*{When $c$ and $f$ are both highly inexplicable}
In the setting, we consider a case where $W_c = 0.1$ and $W_f = 0.05$; this corresponds to the setting where the condition in Lemma \threshref{} does not hold. In this setting we notice that the monotonic trend of the loss always penalizing higher confidence classification to the semantically further class $c$ does not hold. This is because assigning the difference in loss to an almost similar inexplicable class $f$ goes not help in making the classifier more explicable. Yet, similar to the last setting, if the difference is assigned to the correct class the loss penalized the misclassification less, the trend observed as we move from the left-most to the right-most figure.

\subsubsection*{When $c$ and $f$ are both highly inexplicable}
In the setting, we consider a case where $W_c = 0.1$ and $W_f = 0.05$; this corresponds to the setting where the condition in Lemma \threshref does not hold. In \autoref{fig:my_label} (middle), we notice that the monotonic trend of the loss always penalizing higher confidence classification to the semantically further class $c$ does not hold. This is because assigning the difference in loss to an almost similar inexplicable class $f$ goes not help in making the classifier more explicable. Yet, similar to the last setting, if the difference is assigned to the correct class the loss penalized the misclassification less, the trend observed as we move from the left-most to the right-most figure.

\subsubsection*{When $c$ and $f$ are both highly explicable} In the setting, we consider a case where $W_c = 0.4$ and $W_f = 0.5$; this corresponds to the setting where the condition in Lemma \threshref does not hold. Similar to the previous setting where both classes are almost equally inexplicable, the general trend of the loss function values violating the monotonicity of penalizing the loss on placing higher confidence in $f$ holds, as shown in \autoref{fig:my_label} (bottom). Interestingly, even as the probability assigned to the correct class increases, the classifier does not penalize mis-classifications as all misclassifications are highly explicable. In settings where no misclassification will cause a surprise for the human, the use of the vanilla CCE loss forces the classifier to push for the correct class more than our proposed loss function.

\section*{Additional experimental results}

\paragraph{Grad-CAM}
We further assess the performance of our ResNet-EKL model on ImageNet using Grad-CAM attention maps with randomly sampled imagery. We choose EKL due to the hierarchically similar structure between ImageNet-1000 and WordNet, which allows us to compile the semantic distances between each pair of classes without human labelling intervention. In \autoref{fig:imagenet-saliency} we examine the performance of both classifiers which aligns with our previous findings on CIFAR-10 experiments; we observe that when both are correct (images 11--15), ResNet-EKL focuses in smaller regions for its predictions. In image number $12$, the class represents a {\tt mousetrap} and the saliency region is significantly reduced to only fit the class object. This can also be seen in $11$ where the highlighed region for {\tt tripod} does not cover the human in the picture as opposed to the vanilla classifier. When both classifiers are incorrect (images 1--5) we notice that EKL still focuses on semantically-closer regions to the true prediction. Image $1$ shows a {\tt garter snake} misclassified by both for a {\tt sea snake} and while the vanilla classifier focuses to the background, EKL does so to the animal. Further, for image $3$, an {\tt envelope} is misclassified to a {\tt binder} by EKL in comparison to a {\tt laptop} by the base predictor. We notice again that EKL highlights parts of the object itself while the vanilla still gives more importance to the background. Finally, images 6--10 show cases where EKL is right while the base is not. We observe that, as expected, EKL directs its attention directly to the object and, while the base also does in some cases ($7$ and $9$), it fails to correctly label it. In $6$ the {\tt seashore} is directly highlighted by EKL, while the vanilla focuses on the toy confusing it for a {\tt coffeepot}.

\paragraph{Discussion on losses as external evaluation metrics}
In Peterson et al. 2019 authors make extensive use of loss function values as an interchangeable external evaluation metric with accuracy. We posit and show below that the use of loss function values for our explicability needs more experimentation and may not be in direct correlation with actual human-evaluated and our synthetically-defined explicable scores. 

In table 1 of ths appendix we compile the four test losses (CCE, IHL, CHL and EKL) of all CIFAR-10-trained ResNet-v2 classifiers. Rows represent the model trained on each loss while columns represent loss functions. As an example, the value at the top left shows the loss obtained when the vanilla model is evaluated with the $\mathcal{L}_{IHL}$ function. Models trained on a particular loss will tend to minimize that loss better and this can be seen in the diagonal of the matrix. CIFAR-10+ scores show CHL (with $\mathcal{L}_{CHL}$) as the model of choice for CIFAR-10+, however, human experimentations give CHL the lowest explicability value (1.725) showing not a correspondence. Further, CIFAR-10+ average loss value of the three models (diagonal) is roughly the same as CIFAR-10 while the former is certainly less explicable due to its out-of-distribution nature.
Thus, the use of loss function values as explicability metrics prove not to be aligned with human's opinion and, even though more experiments could be performed on this front, we decided to make our own metrics and continue with human evaluations for explicability matters.

\begin{table}[t]
\small
\centering
\begin{tabular}{lcccc}
\toprule
& \multicolumn{3}{c}{\em Loss values} &\\
Model & $\mathcal{L}_{IHL}$ $\downarrow$ & $\mathcal{L}_{CHL}$ $\downarrow$ & $\mathcal{L}_{EKL}$ $\downarrow$ &\\
\midrule
CIFAR-10\\
\hspace{5px}ResNet-v2 (CCE) & $14.76$ & $ 5.04$ & $16.04$\\
\hspace{5px}ResNet-v2 (IHL) &  \textbf{2.25} &  $1.88$ &  $2.31$ \\
\hspace{5px}ResNet-v2 (CHL)  & $3.05$ & \textbf{1.30} & $3.27$ \\
\hspace{5px}ResNet-v2 (EKL) &  $2.35$ & $1.56$ & \textbf{2.46} \\ 
CIFAR-10+\\
\hspace{5px}ResNet-v2 (CCE) & $8.25$ & $ 4.60$ & $8.71$\\
\hspace{5px}ResNet-v2 (IHL) &  \textbf{2.29} &  $2.13$ & 2.31 \\
\hspace{5px}ResNet-v2 (CHL)  & $2.99$ & \textbf{2.06} & $3.09$ \\
\hspace{5px}ResNet-v2 (EKL) &  $2.41$ & 2.02 & \textbf{2.46} \\
\bottomrule
\end{tabular}
\caption{Loss function values from CIFAR-10 trained ResNet-v2 models.}
\label{tab:cifar10-losses}
\end{table}
